\documentclass[conference]{IEEEtran}
\IEEEoverridecommandlockouts

\usepackage{cite}
\usepackage{amsmath,amssymb,amsfonts}
\usepackage{algorithmic}
\usepackage{graphicx}
\usepackage{textcomp}
\usepackage{xcolor}
\usepackage{multirow}
\usepackage{hyperref}
\usepackage{makecell}
\usepackage{float}
\usepackage{xcolor} 

\newcommand{\hejer}[1]{{\color{black}#1}}
\newcommand{\mecdg}[1]{{\color{black}#1}}

\def\BibTeX{{\rm B\kern-.05em{\sc i\kern-.025em b}\kern-.08em
    T\kern-.1667em\lower.7ex\hbox{E}\kern-.125emX}}
\begin{document}

\newcommand\benchm{\textit{WarNav}} 
\newcommand{\comment}[1]{\textcolor{red}{#1}}

\title{\benchm: An Autonomous Driving Benchmark for Segmentation of Navigable Zones in War Scenes
}

%
%
%
%
%
%

\author{
Marc-Emmanuel Coupvent des Graviers$^{1}$,
Hejer Ammar$^{2}$,
Christophe Guettier$^{1}$,
Yann Dumortier$^{1}$,
Romaric Audigier$^{2}$ \\[1ex]
$^{1}$ Safran Electronics and Defense, Massy, France \\
\{marc-emmanuel.des-graviers, christophe.guettier, yann.dumortier\}@safrangroup.com \\[1ex]
$^{2}$ Université Paris-Saclay, CEA, List, F-91120 Palaiseau, France \\
\{hejer.ammar, romaric.audigier\}@cea.fr
}

\maketitle

\begin{abstract} 

We introduce \benchm, a novel real-world dataset constructed from images of the open-source DATTALION 
repository, specifically tailored to enable the development and benchmarking of semantic segmentation models for autonomous ground vehicle navigation in unstructured, conflict-affected environments. This dataset addresses a critical gap between conventional urban driving resources and the unique operational scenarios encountered by unmanned systems in hazardous and damaged war-zones. We detail the methodological challenges encountered, ranging from data heterogeneity to ethical considerations, providing guidance for future efforts that target extreme operational contexts. 
To establish performance references, we report baseline results on \benchm\ using several state-of-the-art semantic segmentation models trained on structured urban scenes. 
We further analyse the impact of training data environments 
and propose a first step towards effective 
navigability in 
challenging  \hejer{environments with the constraint of having no annotation of the targeted images}.
Our goal is to foster impactful research that enhances the robustness and safety of autonomous vehicles in 
high-risk scenarios while being frugal in \hejer{annotated} data.

\end{abstract}

\begin{IEEEkeywords}
Dataset - Annotation - Semantic Segmentation - Unstructured Environments - Navigability - Data frugality.
\end{IEEEkeywords}

\section{Introduction} 

Modern warfare presents significant challenges for the tactical mobility of mounted combat vehicles. Due to contested environments (GPS-denied, RF-denied), vehicles such as battle tanks, infantry fighting vehicles, and autonomous robots cannot rely on outdated operational pictures to achieve mission objectives. Intensive indirect fire rapidly alters navigable space and key-terrain positions, affecting mission feasibility. Tactical missions now require tight integration of situation awareness and \textit{just-in-time} planning. Furthermore, dominant threats (e.g., loitering ammunitions, short loops between UAV and artillery, remote navigation of drones and robots, or improvised explosive devices) further limit navigable space.

These challenges, rooted in the dynamic nature of the battlefield and the diversity of threats, reveal critical limitations in current mobility and navigation systems. While autonomous navigation technologies in modern urban scenes have been widely developed with rich perception modules owing to finely annotated semantic segmentation datasets, their applicability in hostile, unstructured, and destructed combat zones remains highly constrained. In fact, in these situations, robot autonomy or driver assistance will require strong advancements to navigate efficiently in the no man's land. Moreover, due to the lack of geometrically structured shapes, the differences between two scenes are difficult to assess, even by a human expert, and limited dataset is available to reasonably master the learning bias. 

A partial workaround to data scarcity consists of leveraging publicly available information, through techniques such as web scraping, to gain additional information on the target environment. However, the incorporation of extra-military data 
introduces additional risks \cite{rettore_military_2023}. In particular, publicly accessible sources may be subject to intentional manipulation, including large-scale image tampering or disinformation campaigns \cite{alkhowaiter_detecting_2023}.


In this paper, we propose \benchm, a war-zone-specific dataset constructed from the DATTALION repository~\cite{dattalion} to support the development and evaluation of robust semantic segmentation models for 
navigability purposes 
in 
conflict-affected settings. The goal 
is to bridge the domain gap between traditional urban driving datasets and the operational realities faced by unmanned systems in hazardous areas. The central challenges lie in 
collecting, filtering, annotating, and validating imagery that is both representative and ethically sourced, while establishing procedures that ensure the resulting dataset meets the rigorous standards required for both academic research and practical deployment. Several techniques have been applied to meet these criteria for \benchm. Indeed, semantic class labels tailored to navigation tasks are proposed for the test and validation sets to enable performance evaluation.


Moreover, we report baseline performances of several models trained on available annotated datasets 
without any exposure to \benchm\ images. Test and validation sets are used to evaluate them in war-zone challenging regions, by varying the model architectures, the backbones, and the memory footprints. We also assess the impact of training data domains, ranging from urban to rural and from structured to less-structured environments, on segmentation effectiveness. Results highlight that each domain offers unique benefits towards robust navigability in destructed outdoor areas. Finally, we propose a simple yet effective frugal approach that delivers strong perception capabilities under resource constraints.

Our contributions can be summarized as follows:
\begin{itemize}
    \item We introduce a novel and challenging use case for semantic segmentation in war-damaged environments, targeting frugal autonomous navigation.
    \item We construct the \benchm\ dataset via a pipeline of image selection, filtering, curation, and annotation, with a strong focus on ethical sourcing, providing practical insights for future dataset design in extreme deployment scenarios.
    \item We provide performance on \benchm\ of diverse baselines by varying models, backbones or training environments, and propose an initial frugal approach achieving 
    effective navigability segmentation in conflict-affected areas.
\end{itemize}


    
    


\section{Autonomous Robot Use case presentation} 

\subsection{Goal and challenges}

The advance of autonomous and assisted driving technologies is highly dependent on the availability of extensive, high-quality datasets for model development and validation. However, most of the existing datasets for semantic segmentation in the context of ground vehicles, such as Cityscapes \cite{Cityscapes} or KITTI \cite{kitti}, are predominantly collected in highly structured and undisturbed urban environments. This limits their relevance and utility when models are deployed in more complex, degraded, or unstructured real-world contexts. Through this use case, our aim is to contribute not only with a valuable data resource for the research community but also methodological guidance for future efforts in dataset construction for extreme or atypical operational contexts.

\subsection{Semantic Segmentation of Navigable Spaces}

One particularly challenging use case arises in the domain of military operations, where unmanned ground vehicles (UGVs) are expected to perform autonomous navigation tasks in environments characterised by significant destruction, involving debris, destructed vehicles, shell holes, ruts, collapse of buildings, or landslides. In such contexts, accurate perception is critical for both navigation effectiveness and safety. Specifically, the characterization of drivable areas with obstacles can be improved using semantic segmentation. Thanks to semantic retrievals, on-board planners can provide navigation instructions (maneuvers, paths, trajectories) for automatic path and mission completion. However, data scarcity is a major limitation: operational constraints and safety concerns make it impractical to acquire and exhaustively annotate large-scale, representative image datasets in these environments.

\subsection{Frugality needs for autonomous navigation with local situation awareness}

Autonomous driving in complex, destructured or unstructured environment must be robust to changes. In particular for ground robotic, mission planning and execution must account for the ability of the autonomous system to interpret its environment, using semantic segmentation among other mission information available on board \cite{Guettier_Lucas_2016}. \mecdg{Moreover, typical deployment of robotics in military context implies late in-situ image acquisition. It thus can rely on model adaptation during mission preparation \cite{Guettier_Lamal_Mayk_Yelloz_2015} through three main phases:}
\begin{itemize}
\item At mission preparation time, where rough data terrain are available, but not necessarily representative of the battlespace environment.
\item After the first mission execution, where some sparse data are gathered from the executed navigation plan. This would correspond to a first major model adaptation.
\item During repetitive mission operations, where incremental model adaptations could be performed thanks to incremental data retrieval.
\end{itemize}

\subsection{Providing dataset from conflict zones}

To address this challenge, we turn to publicly available resources that offer authentic, situationally relevant visual content. The DATTALION repository \cite{dattalion} is a prominent example, providing visual documentation from Ukrainian conflict zones, reflecting the diversity and chaos of post-conflict urban environments. However, directly leveraging such open-source imagery for machine learning applications presents several challenges. The imagery is heterogeneous in terms of scene content and neither curated nor annotated for technical use cases such as semantic segmentation. Furthermore, issues of data privacy and ethical use must be rigorously addressed when dealing with potentially sensitive imagery featuring vulnerable civilians or recognisable features.

\section{\benchm: a benchmark for frugal segmentation of navigable zones in war scenes}

\subsection{DATTALION: a dataset of real war scene images} 

The DATTALION dataset \cite{dattalion} is a large open-source multimedia repository documenting the Russian invasion of Ukraine, launched in 2022. It consists of over 4,000 verified videos and 20,000 images, along with metadata including location, date, source, and type of event (e.g., attacks on civilian infrastructure, troop movements). The dataset is maintained by a volunteer-driven Ukrainian initiative and is primarily intended to support research, journalism, and accountability efforts related to war crimes and conflict analysis. The dataset is organized chronologically with monthly chunks.
For autonomous vehicle research, only a subset of DATTALION is relevant. Many images, such as indoor scenes, nighttime photographs, or close-ups, do not provide useful information for training perception systems designed for drivable area segmentation in outdoor \hejer{daytime} environments.


\subsection{Image Selection} 

We have first performed an initial assessment of the suitability of the DATTALION content for autonomous navigation zone detection. We have found multiple examples of outdoor road areas with partially damaged buildings or vehicles. We also found interesting scenarios such as crop field wildfires or road blast craters, which would be particularly difficult to recreate if we had to design a testing area for new image acquisition.  

We then performed a progressive filtering and selection process. This filtering approach is based on past experience in artistic image competitions \footnote{\mecdg{https://www.salondaguerre.paris/}} 
where image quality assessment is typically performed in a few seconds during the first selection rounds. This experience has shown that selecting a few thousand images from a pre-existing repository is feasible in a reasonable time by a small dedicated team. \mecdg{The use of automated image preselection, such as Vision Language Models, was not considered so far, as their robustness in destructured environment was unknown.}

The following methodological steps were undertaken:

\begin{itemize}
    \item Submission of a data processing declaration in accordance with the General Data Protection Regulation (GDPR), specifying the use of encrypted hard drives and the deletion of image data upon completion of the selection process.
    \item Downloading of the DATTALION dataset, retaining only image files for analysis. All video files and Word documents were excluded from further consideration.
    \item Development of a standardized image selection protocol, including representative examples of images to be retained or discarded, based on relevance to research objectives and image quality.
    \item Initial filtering of the dataset through exclusion of images based on the following criteria: nighttime scenes, close-up object views, indoor settings and building facades without visible road infrastructure \hejer{as only outdoor daytime scenes are relevant for our use case. Images containing blood, cadavers, or partial blurring were also removed for ethical and bias considerations.}. This filtering process was conducted in parallel by team members, each responsible for a designated subset of monthly data.
    \item Manual review of the pre-filtered images to remove remaining outliers. This step was significantly faster than the initial filtering, thanks to the reduced volume of images requiring inspection.
    \item Partitioning of the monthly image subsets into training (5354 images from 8 months), validation (100 images from one month), and testing (100 images from 2 months) datasets. Note that there is no overlap between the months represented in the three sets to avoid domain leakage.
\end{itemize}

It is worth noting that several original images from the DATTALION dataset are partially blurred. These blurred regions typically correspond to cadavers or individuals whose identities were likely intentionally obscured for privacy or ethical reasons. To avoid introducing a potential bias during training, where a semantic segmentation model might learn to associate blurring artifacts with the presence of persons, we opted to discard such images. Conversely, images containing unblurred yet unidentifiable individuals were retained without modification, under the assumption that they resemble data that could be passively captured by onboard cameras of autonomous vehicles.

\subsection{Semantic Classes} \label{classes} 

Based on the intended use case and the availability of this rich dataset, the set of semantic classes to be annotated was progressively refined. The following definitions were ultimately adopted:

\begin{itemize}
    \item \textbf{Overlay}: Regions containing graphical overlays or annotations that were added post-capture. These pixels are excluded from both training and performance evaluation, as they do not correspond to real-world scene content.
    \item \textbf{Road}: Surfaces intended for civilian vehicular traffic, typically paved with asphalt or similar materials.
    \item \textbf{Drivable}: Areas that are not formal roads but are deemed traversable by military 4x4 vehicles (e.g., dirt paths, open fields).
    \item \textbf{Pedestrian}: Humans. Accurate detection of this class is essential for tasks related to safe autonomous navigation.
    \item \textbf{Vehicle}: Civilian vehicles that are potentially operable. Obstacle avoidance algorithms would consider them as potentially non-static obstacles. Damaged or abandoned car wrecks are excluded from this category.
    \item \textbf{Background}: All 
    remaining 
    regions are classified as background, encompassing areas where navigation is not feasible (e.g., buildings, vegetation, sky, rubble, blast craters or other static obstacles).
\end{itemize}


\subsection{Annotation} \label{sec:annotation} 


Even if unsupervised techniques are foreseen to address annotation constraints, pixel annotation is necessary for performance evaluation. This annotation is performed only on validation (val) and test sets. The training dataset remains completely unannotated to emphasize the need for unsupervised learning strategies suited to real-world constraints. In practice, less than 4\% (i.e., 200 among 5554) of selected images were annotated.

The annotation process began with an initial calibration phase during which a small sample of images was annotated and then discussed to clarify expectations and resolve ambiguities. The following annotation guidelines were established and agreed upon:

\begin{itemize}
    \item \textbf{Annotation method:} Semantic segmentation was performed by manually outlining regions of interest using polygons. Each segmented pixel is assigned to exactly one semantic class; no overlapping segments.
    \item \textbf{Obstacle annotation:} Small debris or wreckage that could realistically be traversed by a military vehicle were not annotated individually. Conversely, blast craters are generally considered non-drivable and should be explicitly labelled as \texttt{background}.
    \item \textbf{Surface transitions:} Border zones between different drivable surfaces—such as the interface between asphalt and cobblestone or between paved and unpaved areas—are to be labelled as drivable if they are visually and functionally navigable.
    \item \textbf{Occluded road surfaces:} When dense vegetation completely obscures the underlying ground, the surface condition cannot be reliably assessed. In such cases, the region must be labelled as \texttt{background}, as no inference should be made without clear visual evidence.
    \item \textbf{Sparse foreground elements:} Objects such as tree branches, leaves, or overhead cables, which do not obstruct vehicle motion but may appear in the foreground, are not annotated.
    \item \textbf{Vehicle versus static obstacle distinction:} The boundary between a functional vehicle and an immobile obstacle can be ambiguous, especially in war-zone imagery. The chosen criterion is based on potential operability: only vehicles that appear to be intact and potentially capable of movement are labelled as \texttt{vehicle}. Severely damaged vehicles (e.g., burned-out shells, or dismembered car halves) are treated as part of the \texttt{background}.
\end{itemize}


All test and validation images were manually annotated following this protocol. The resulting annotation masks were saved using the Cityscapes file format \cite{cityscapes_format}. 

To assess the consistency and reliability of human annotation, a subset of 10 images from the test set was independently annotated by two additional annotators, resulting in three distinct annotations per image. 
The inter-annotator agreement was evaluated on all pixels: $92.3\%$ of them were assigned identical labels by all three annotators, indicating a high level of consistency. However, $7.7\%$ pixels showed at least one disagreement and only $0.17\%$ pixels were assigned three completely different labels, reflecting localised interpretation ambiguities. The mean pixel-wise entropy in the dataset was relatively low ($0.0492$), further supporting strong annotation consistency. Pairwise Dice similarity coefficients were calculated between annotators for each semantic class. High agreement was observed in classes such as \texttt{background}, \texttt{vehicles}, \texttt{overlay} and \texttt{pedestrian} with Dice scores exceeding $0.95$ across all annotator pairs. Moderate discrepancies appeared in \texttt{drivable} and \texttt{road} classes, which yielded lower Dice scores. In fact, these classes may be more prone to subjective interpretation or boundary ambiguity due to their close definitions (i.e., zones drivable by a civilian car vs. a 4x4 military vehicle). Nonetheless, these inconsistencies are not critical for the intended military application, as all affected areas still fall within the broader category of navigable space which is our primary concern. The inter-annotator agreement from this sample will serve as a benchmark for evaluating the performance of automated semantic segmentation models. Otherwise stated, we will consider the annotations having the smaller discrepancy with the two others (i.e., \textit{Annotator 2}).

\begin{figure}[htbp]
  \centering
  \includegraphics[width=0.9\columnwidth]{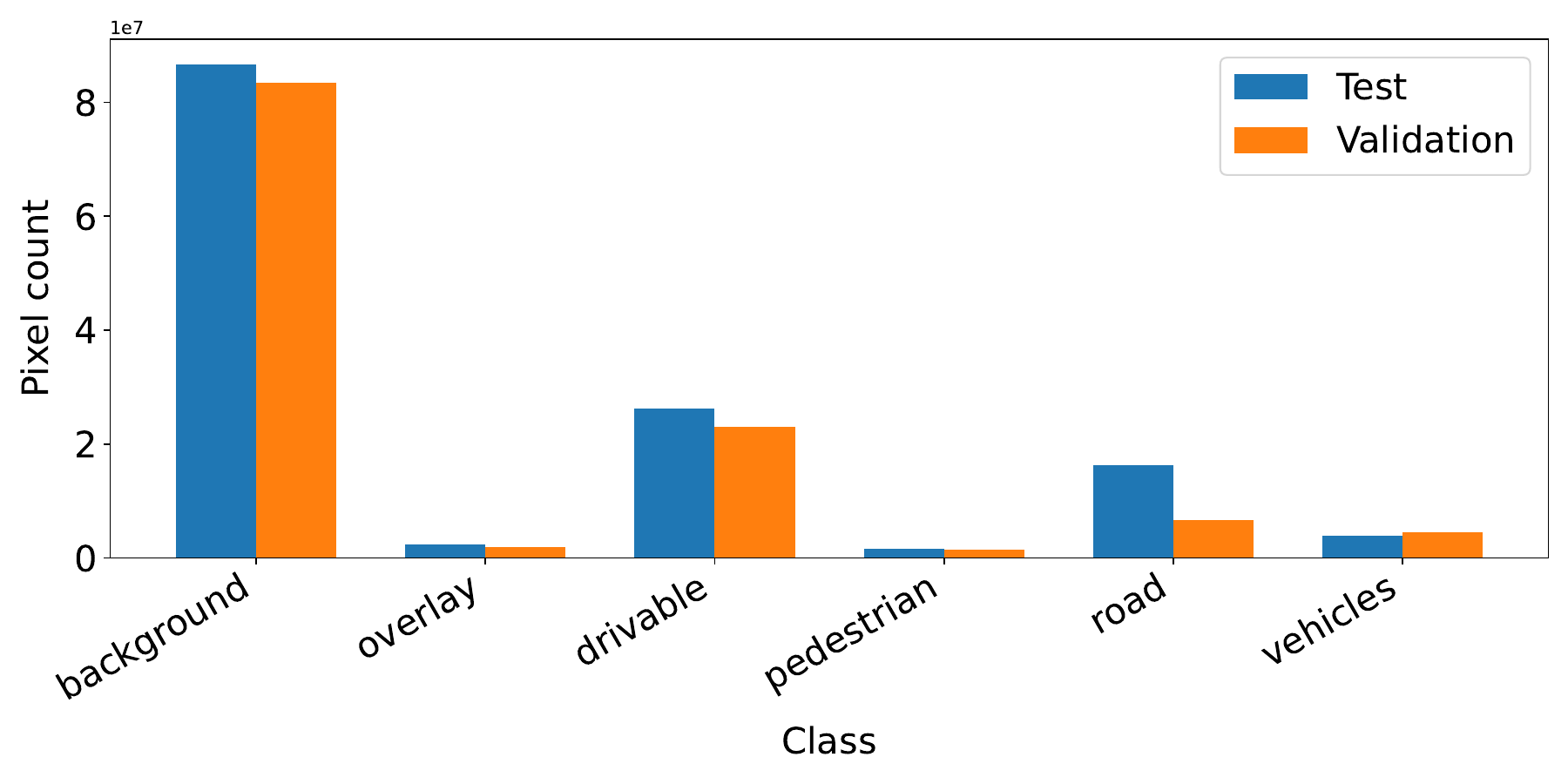}
  \vspace{-4mm}
  \caption{Histogram of number ($\times 10^7$) of pixels per class for the test and the validation sets of \benchm. When ignoring `overlay', the 5 remaining classes constitute the so-called $L_5$ setting used in this paper.}
  \label{fig:pixels}
\end{figure}

\begin{figure}[htbp]
  \centering
  \includegraphics[width=0.9\columnwidth]{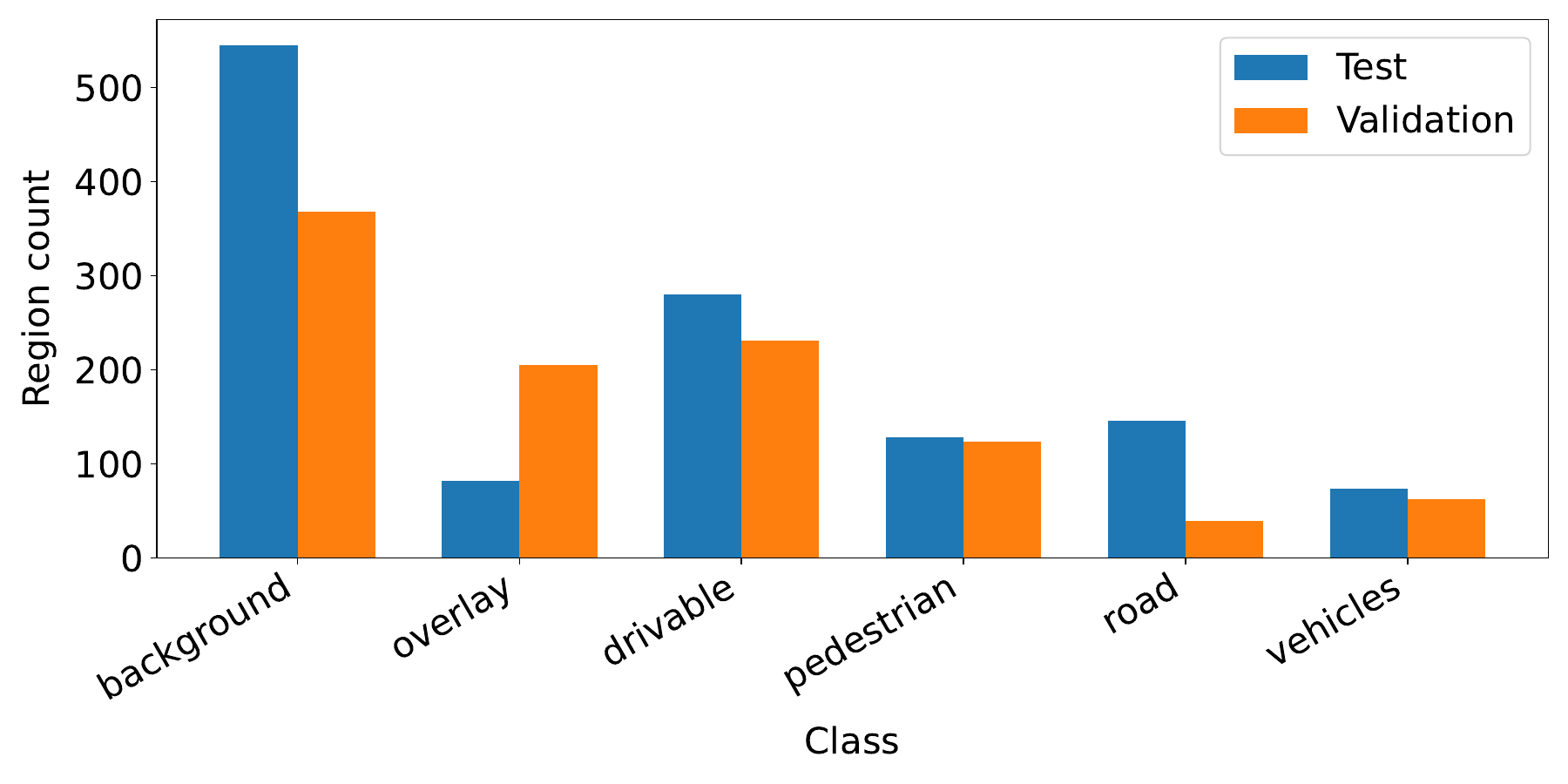}
   \vspace{-4mm}
  \caption{Histogram of connected regions per class for the test and the validation sets of \benchm.}
  \label{fig:regions}
\end{figure}

Figure~\ref{fig:pixels} illustrates the distribution of pixel classes showing a strong dominance of the background class, followed by drivable areas and roads, which together account for the majority of labelled pixels. In contrast, pedestrian and vehicle classes appear significantly less frequently, which is predictable due to the war context and to their smaller size. Figure~\ref{fig:regions} illustrates the region count histogram providing insight into the spatial distribution and fragmentation of each class. While background regions remain dominant, classes like pedestrian and vehicles exhibit a higher number of small, disconnected regions relative to their pixel count. The similarity in distributions between the test and validation sets in both histograms indicates good consistency in annotation quality and dataset structure, which is crucial for reliable performance assessment.

\vspace{-4mm}

\vspace{-4mm}

\subsection{Dataset Open-sourcing}
Selected images and annotations are available on \url{https://github.com/CEA-LIST/WarNav}. It provides DATTALION image names for the different splits and annotation masks for test and validation datasets. \hejer{The original images are not shared due to licensing restrictions.}

\begin{table*}[h]
\renewcommand{\arraystretch}{1} 
\centering
\begin{tabular}{llrcccc}
\hline
\multirow{2}{*}{Architecture} & \multirow{2}{*}{Backbone} & \multirow{2}{*}{\#P(M)} & \multicolumn{2}{c}{mIoU (in \%)} & \multicolumn{2}{c}{wmIoU (in \%)} \\
\cline{4-7}
& & & Cityscapes(val,$L_{19}$) & Cityscapes(val,$L_{5}$) & \benchm (test,$L_{5}$) & \benchm (val,$L_{5}$) \\
\hline
DeepLabv3+ \cite{deeplabv3+} & ResNet101 \cite{resnet} & 66 & 76.2 & 91.2 & 53.3 & 46.7 \\
\hline
Mask2Former \cite{mask2former} & SwinB \cite{swin} & 104 & \textbf{83.3} & \textbf{93.5} & 51.4 & 49.8\\
\hline
SegFormer \cite{segformer} & MiT-B5 \cite{segformer} & 85 & 82.4 & 92.7 & \textbf{61.5} & \textbf{58.1} \\
\hline
\end{tabular}
\caption{Performances of different approaches based on different backbones all trained on the Cityscapes train-set. For each method, we provide the number of parameters in millions (\#P(M)), mIoU results on Cityscapes val-set considering the $L_{19}$ and $L_{5}$ labels settings, and the wmIoU results on the \benchm\ test and val sets. Best results per column are in \textbf{bold}.} 
\vspace{-5mm}
\label{tab:approaches}
\end{table*}

\section{\hejer{Frugal} baselines for \benchm\ benchmark} 
\subsection{wmIoU: A new weighted mIoU suitable for \benchm} \label{wmiou} 
Although the mean Intersection over Union (mIoU) is the standard metric for evaluating semantic segmentation performance, it may obscure critical aspects relevant to our specific use case as it equally considers all pixels. First, since our primary goal is to ensure reliable navigability, we place greater importance on accurately segmenting regions closer to the vehicle 
than on distant areas. This distinction is particularly significant for the \texttt{background} class, as it encompasses both navigational obstacles such as rubble and debris, and other non-navigable regions such as sky and buildings. In our context, identifying obstacles within navigable zones is more crucial than segmenting other \texttt{background} elements, as they have a more immediate impact on navigation decisions. Secondly, we argue that accurately segmenting 
the inner parts of each zone is more critical than precisely delineating contours, particularly at the boundaries between \texttt{road} and \texttt{drivable} areas. To reflect these priorities, we propose a new weighted mIoU (wmIoU) that accounts for both factors, \hejer{by weighting the ground truth class label map $C_{gt}$ with a weight map $W$ using a Hadamard product \cite{hadamard} (here denoted $\circ$) such as proposed by \cite{wIoU}:}
\begin{equation}
    wIoU = \frac{ 
    | (C \cap C_{gt}) \circ W| }
    {| (C \cup C_{gt} ) \circ W |} \label{wiou}
\end{equation}
where $C$ denotes the predicted class label map. Note that the final $wmIoU$ score is obtained by averaging the $wIoU$ values across all classes.


We draw inspiration from this work and adapt it to align with our objectives. Specifically, we introduce a distance map $D=D_1 \circ  D_2$ which incorporates our two criteria:
\begin{itemize}
    \item We consider the highest non-background pixel $p_{fg}$ as the horizontal limit between the most critical regions that contain navigable zones (below) and the less relevant non-navigable areas (above). To reflect this distinction, we construct $D_1$ as \hejer{two piecewise decreasing linear functions}
    $f(a,b)$ defined by their extrema $a$, $b$,
    assigning greater weights to closer pixels and especially the more critical foreground ones (i.e., satisfying $p$ below $p_{fg}$):
    \begin{center}
\begin{equation}
        D_1(p)=\left\{ 
    \begin{array}{ll}
    f((0,1),(p_{fg},0.8)) & \text{if } p \text{ below } p_{fg} \\
    f((p_{fg},0.2),(p_{max}, 0.1))  & \text{otherwise}
    \end{array}
    \right.
\end{equation}
\end{center}
\item We compute a boundary distance (a.k.a. distance transform) map $D_2$, \hejer{where for each pixel $p$, $D_2(p)$ is the minimum distance to a pixel
of a different class
normalized by the maximum value found in its connected component}.
\end{itemize}
The resulting map $D$ is then used to create a weight map $W(p)=e^{\alpha D(p)}$, to compute a $wIoU$ per class such as presented in Eq.~\ref{wiou}, \hejer{where $\alpha=0.3$ controls the slope decay}. This formulation accentuates regions farther from class boundaries, prioritizes forefront areas, closer to the camera, and especially emphasizes foreground pixels. Thus, the influence of distant background regions, which often dominate the image but are less relevant for immediate navigation, is reduced.

\subsection{Datasets} 
In addition to \benchm, we consider three public datasets:

\textbf{Cityscapes \cite{Cityscapes}} is a commonly used dataset for semantic segmentation for autonomous driving. It contains $2975$ finely annotated training images, and $500$ validation images (val), all segmented into 19 semantic classes: $L_{19}$. Notably, the dataset mainly features scenes from well-structured urban environments, representing organised and structured cities.

\textbf{RUGD \cite{RUGD}} is a video dataset captured in rural and less structured outdoor environments,  offering more representative samples for complex rural scenes. The original dataset is divided into $4759$ train, $733$ validation and $1964$ test images. We modify this split to $4375$ for training, $1240$ for validation (val), and $1841$ for testing, to (i) reduce the size of the training set for better comparability with the Cityscapes setup, (ii) ensure the inclusion of the class \texttt{water} in the training set, and (iii) minimize domain leakage across splits. The images are annotated into $24$ possible class labels.

\textbf{Earthquake-site database \cite{earthquake}} (referred to as `\textit{Earthquake}' in this paper) is a set of images  depicting earthquake-related damage. It was finely segmented into $10$ semantic classes such as every small crack, wreckage, or obstacle is highlighted, in contrast to \benchm\ where only bigger obstacles or blast craters obstructing military vehicle motion are considered. 
This dataset includes scenes of both urban and rural environments, with $686$ train and $50$ test images.

\subsection{Experiments and results} 
In this section, we provide several baseline performances on the test and val sets of \benchm, analysing the influence of model architecture, backbone size, and training dataset. It should be noted that none of the models used were trained using images from \benchm. Instead, we report inference results from models trained on public annotated datasets. Indeed, there is an important domain gap between these datasets and \benchm. The presented results serve as initial baselines and provide insights into how various model characteristics influence performance in our specific application setting. \hejer{Publicly available chekpoints were used to produce the results in Tables \ref{tab:approaches} and \ref{tab:backbones}. For Table \ref{datasets}, we employed the official SegFormer code \cite{segformer}, with minor modifications to the dataloaders to accommodate the different datasets.}

\begin{table*}[h]
\renewcommand{\arraystretch}{1} 
\centering
\begin{tabular}{crcccc}
\hline
\multirow{2}{*}{Backbone} & \multirow{2}{*}{\#P(M)} 
& \multicolumn{2}{c}{mIoU (in \%)} 
& \multicolumn{2}{c}{wmIoU (in \%)} \\
\cline{3-6}
& & Cityscapes(val,$L_{19}$) & Cityscapes(val,$L_{5}$) & \benchm (test,$L_{5}$) & \benchm (val,$L_{5}$) \\
\hline
MiT-B0 & 3.7 & 76.3 & 90.8 & 56.0 & 52.3 \\
\hline
MiT-B1 & 13.7 & 78.5 & 91.8 & 54.9 & 49.8 \\
\hline
MiT-B2 & 27.5 & 81.0 & 92.4 & 55.6 & 53.2 \\
\hline
MiT-B3 & 47.3 & 81.7 & \textbf{92.7} & 58.9 & 55.2 \\
\hline
MiT-B4 & 64.1 & \textbf{82.7} & \textbf{92.7} & 60.6 & 56.4 \\
\hline
MiT-B5 & 84.7 & 82.4 & \textbf{92.7} & \textbf{61.5} & \textbf{58.1} \\
\hline
\end{tabular}
\caption{Performances of SegFormer \cite{segformer} based on different backbones all trained on the Cityscapes train-set. For each model we provide the number of parameters in million (\#P(M)), \textit{mIoU} results on Cityscapes \textit{val} set considering both $L_{19}$ and $L_{5}$ label settings, and the \textit{wmIoU} results on the \benchm\ \textit{test} and \textit{val} sets. Best results per column are in \textbf{bold}.} 
\vspace{-5mm}
\label{tab:backbones}
\end{table*}

\textbf{Effect of model architecture:} 
First, we provide in Table~\ref{tab:approaches} a comparison between various state-of-the-art segmentation models all trained on the Cityscapes~\cite{Cityscapes} training set to segment images into 19 possible semantic classes ($L_{19}$). We chose a CNN-based model (i.e., DeepLabv3+~\cite{deeplabv3+}), and two visual transformer-based (ViT~\cite{vit}) approaches usually providing better results: Mask2Former~\cite{mask2former} and SegFormer~\cite{segformer}. These models have different architectures, are based on different backbones (i.e., ResNet101~\cite{resnet}, SwinB~\cite{swin} and MiT-B5~\cite{segformer}), and have different memory footprints (see number of parameters \#P(M) in Table~\ref{tab:approaches}).

For each approach, we report \textit{Cityscapes(val,$L_{19}$)}: the \textit{mIoU} performance on the Cityscapes \textit{val} set segmented into $L_{19}$. These results illustrate the in-domain semantic segmentation performance as both training and evaluation are conducted on subsets of the same dataset with consistent class labels. As anticipated, ViT-based methods significantly outperform DeepLabv3+, with larger model variants achieving higher \textit{mIoU} scores.

Moreover, for a better comparability with the \benchm\ benchmark, we propose to map each class from $L_{19}$ to one of the 5 classes  $L_{5}$ of \benchm\ as follows ($L_{19} \rightarrow L_{5}$):
\begin{itemize}
    \item road $\rightarrow$ road;
    \item sidewalk and terrain $\rightarrow$ drivable;
    \item person and rider $\rightarrow$ pedestrian;
    \item car, motorcycle, bicycle, truck, bus and train $\rightarrow$ vehicle;
    \item sky, vegetation, building, fence, wall, pole, traffic sign and traffic light $\rightarrow$ background.
\end{itemize}
As explained in Sec.~\ref{classes}, we omit the \texttt{overlay} class during evaluation. We apply this mapping to all Cityscapes val prediction and ground truth segmentation maps and perform a new mIoU over the resulting $L_{5}$: Cityscapes(val,$L_{5}$). These values are higher than Cityscapes(val,$L_{19}$) due to the merging effect of fine-grained object classes into broader categories, which simplifies the task. For example, confusion between poles, traffic signs, and traffic lights becomes irrelevant when these are grouped into a single class. Moreover, under this mapping, the performance gap between the three evaluated approaches narrows significantly, with only a $2.3$ p.p. (percentage point) mIoU difference compared to a $7.1$ p.p. gap with the original $L_{19}$ evaluation as even smaller CNN-based models succeed in performing well on this easier task.

The same mapping $L_{19} \rightarrow L_{5}$ is applied to predictions on test and val sets of \benchm, which are compared to the ground truth annotations to compute \benchm (test,$L_{5}$) and \benchm (val,$L_{5}$) respectively, using the \textit{wmIoU} metric. \hejer{In fact, as outlined in Sec.~\ref{wmiou} this metric is more convenient for \benchm\ dataset, contrary to other contexts such as autonomous driving in urban environments}. Interestingly, the lightweight ViT-based segmentation model, SegFormer~\cite{segformer}, achieves the best results on both sets. \hejer{This could be explained by the fact that Mask2Former~\cite{mask2former} is a panoptic segmentation model distinguishing not only the semantic concepts but also individual instances, tending to overfit to specific training instances which reduces generalization in new domains where visual patterns differ.} Thus, we will use SegFormer~\cite{segformer} in the subsequent analyses. 
Note that the gap between the displayed test values and those obtained using different annotations for $10$ images (see Sec.~\ref{sec:annotation} for details) is always less than $0.3$ p.p. \textit{wmIoU}, which confirms the consistency of the annotations.

\begin{table*}[h]
\renewcommand{\arraystretch}{1} 
\centering
\begin{tabular}{lccccc}
\hline
\multirow{2}{*}{Training Data} 
& \multicolumn{3}{c}{mIoU (in \%)} 
& \multicolumn{2}{c}{wIoU (in \%)} \\
\cline{2-6}
& Cityscapes(val,$L_{12}$) & RUGD(test,$L_{12}$) & Earthquake(test,$L_{12}$) & \benchm (test,$L_{5}$) & \benchm (val,$L_{5}$) \\
\hline
Cityscapes \cite{Cityscapes} & \textbf{89.1} & 41.1 & 52.1 & \underline{58.8} & \underline{59.9} \\
\hline
RUGD \cite{RUGD} & 51.3 & \textbf{71.5} & 41.9 & 45.6 & 44.6 \\
\hline
Earthquake \cite{earthquake} & 61.2 & 39.2 & \underline{73.9} & 56.0 & 57.9 \\
\hline
Cityscapes+RUGD+Earthquake & \underline{87.4} & \underline{68.7} & \textbf{75.9} & \textbf{64.9} & \textbf{63.9} \\
\hline
\end{tabular}
\caption{Performance of SegFormer(MiT-B5) \cite{segformer} trained on different datasets. For each model, \textit{mIoU} results on Cityscapes(\textit{val}), RUGD(\textit{test}) and Earthquake(\textit{test}) consider the $L_{12}$ label setting whereas \textit{wIoU} results on the \benchm\ \textit{test} and \textit{val} sets consider the $L_5$ setting. Best results per column are in \textbf{bold}, second best are \underline{underlined}.} 
\vspace{-5mm}
\label{datasets}
\end{table*}

\textbf{Effect of backbone size:} Table~\ref{tab:backbones} presents a comparative analysis of various SegFormer~\cite{segformer} backbones, from MiT-B0 to MiT-B5, in terms of model complexity and segmentation performance with the same evaluation settings. \hejer{More details about the computational costs of each model can be found in \cite{segformer}.} Similarly to Table \ref{tab:approaches}, all models are trained on the Cityscapes \textit{train} set to segment images into $L_{19}$. As expected, increasing the memory footprint leads to improved results, particularly for \textit{Cityscapes(val,$L_{19}$)}, where mIoU rises from $76.3\%$ for MiT-B0 to $82.7\%$ for MiT-B4, with MiT-B5 closely following at $82.4\%$. When evaluating the coarser 5-class $L_5$ setting of Cityscapes, performance differences become less pronounced, with all models achieving scores in a narrow range between $90.8\%$ and $92.7\%$. This confirms our suggestion that collapsing fine-grained categories into broader classes for Cityscapes simplifies the segmentation task, reducing the performance gap between smaller and larger models. 

However, \benchm\ reveals a larger \textit{wmIoU} gap between small and large models driven by the benchmark’s complexity and the domain gap between the structured cities of Cityscapes and the severely damaged environment of \benchm\,. Indeed, \textit{wmIoU} scores gradually improve with model size from $56.0\%$ (MiT-B0) to $61.5\%$ (MiT-B5) for the \textit{test} set, and from $52.3\%$ to $58.1\%$ for the \textit{val} set. Note that results are consistent across \textit{test} and \textit{val} sets for all models, reflecting the reliability of annotations and the representativeness of the selected images for the conflict-affected use case.

\begin{figure*}[h]
  \centering
  \includegraphics[width=1\textwidth]{
  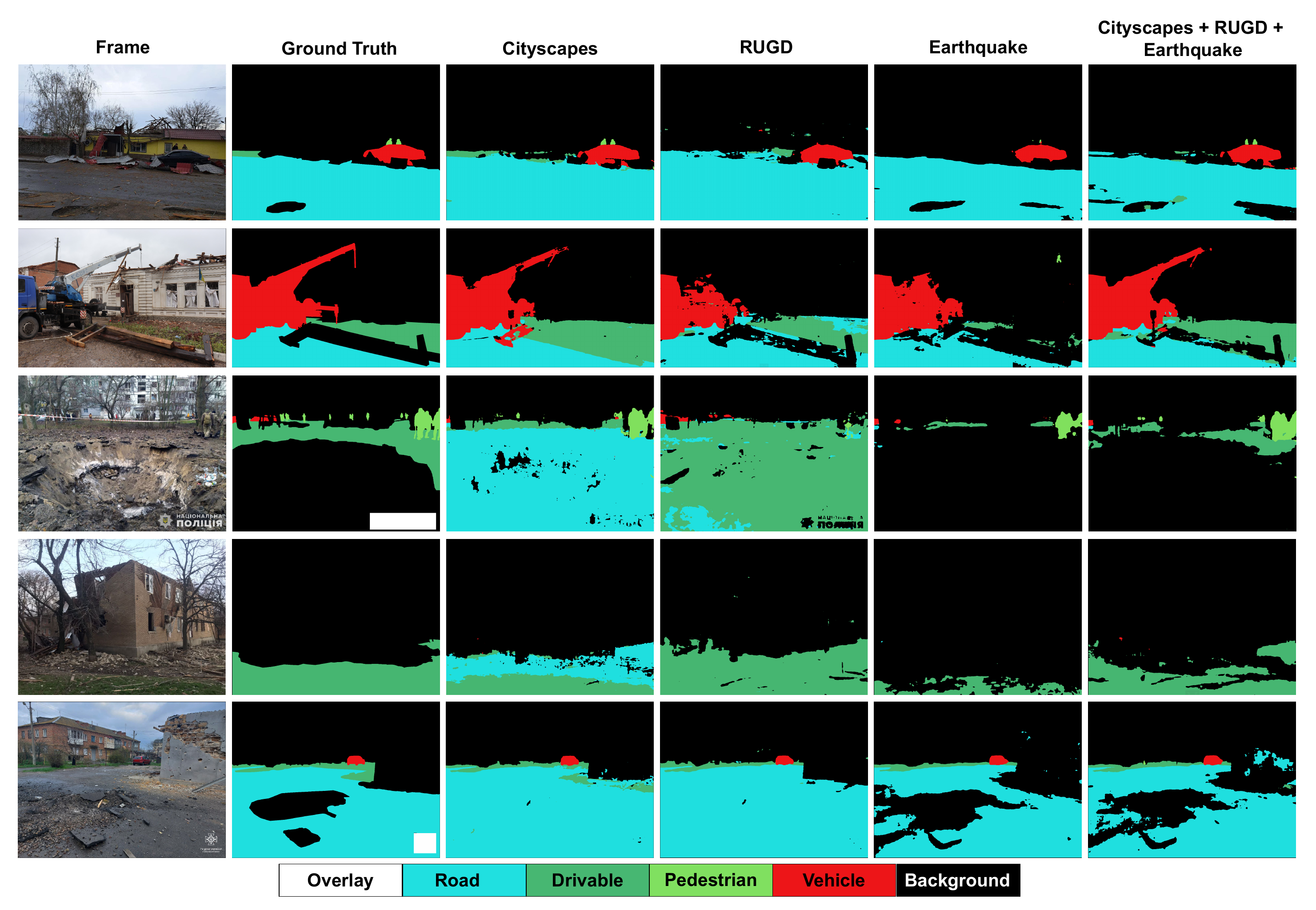} 
  \vspace{-5mm}
  \caption{Illustration of the influence of the training datasets. Columns from left to right are: test images of \benchm, their corresponding annotations, predictions of SegFormer(MIT-B5) trained on Citysapes, RUGD, Earthquake and the combination of the three datasets. }
  \vspace{-5mm}
  \label{fig:predictions}
\end{figure*}

\textbf{Effect of training dataset:} Since Cityscapes primarily features well-structured urban environments, models trained exclusively on Cityscapes often fail to accurately segment destruction-related elements in \benchm\ benchmark (see column 3 in Fig.~\ref{fig:predictions}). 
In this section, we investigate the impact of training data by using different datasets, representing distinct types of outdoor scenes, ranging from structured urban settings to rural and destructed environments.

To ensure a fair comparison between models trained on different datasets, and since each dataset provides its unique class labels and definitions, we introduce a unified label set, $L_{12}$, consisting of 12 high-level semantic categories (super-classes). The labels of each dataset are mapped to this common taxonomy, as detailed in Table~\ref{tab:map}. Specifically, we retain the categories \texttt{road}, \texttt{drivable}, and \texttt{pedestrian} from $L_5$ \benchm, but refine the remaining classes as we believe that combining very distinct semantic concepts during training can harm performances. Thus, the \texttt{vehicle} category is split into three classes: \texttt{car} (civilian cars), \texttt{two wheels} (bicycles and motorcycles), and \texttt{other vehicle} (larger vehicles). The broad \texttt{background} class is further divided into: \texttt{sky}, \texttt{vegetation}, \texttt{buildings}, \texttt{road obstacles} (obstacles located on the roadway), \texttt{side obstacles} (objects found outside the road area), and \texttt{water}.

Note that some inconsistencies were noticed in the annotations of Earthquake. First, grass is inconsistently annotated as either \texttt{vegetation} or \texttt{other}. As a solution, we relabel these areas as \texttt{terrain} when they are predicted as such by the SegFormer(MiT-B5) model trained on Cityscapes $L_{19}$. Second, in the original annotation of Earthquake, all types of vehicles are grouped under a single label. We refine this by using the same model to pseudo-label individual vehicles into: \texttt{car}, \texttt{motorcycle}, \texttt{bicycle}, \texttt{truck}, \texttt{bus}, and \texttt{train}.

\begin{table}[h!]
\renewcommand{\arraystretch}{1} 

\centering
\begin{tabular}{ccccc}
\hline
\textbf{\makecell{$L_{12}$ super-class}} & \textbf{\makecell{Cityscapes}} & \textbf{\makecell{RUGD}} & \textbf{\makecell{Earthquake }} & \textbf{\makecell{\benchm }} \\
\hline
Road     & road  & \makecell{asphalt \\ gravel \\ concrete} & road & road  \\
\hline
Drivable & \makecell{sidewalk \\ terrain} & \makecell{dirt \\ sand \\ grass \\ mulch \\ rockbed} & terrain & drivable \\
\hline
Person & \makecell{person \\ rider} & person & person & pedestrian \\
\hline
Car & car & vehicle & car &  vehicle \\
\hline
Two wheels & \makecell{motocycle \\ bicycle} & bicycle & \makecell{motocycle \\ bicycle} & vehicle \\
\hline
Other vehicle & \makecell{truck \\ bus \\ train} & - & \makecell{truck \\ bus \\ train} & vehicle \\
\hline
Sky & sky & sky & sky & background \\
\hline
Vegetation & vegetation & \makecell{tree \\ bush} & vegetation & background \\
\hline
Buildings & building & \makecell{building \\ bridge} & building  & background  \\
\hline
Road obstacles & - & \makecell{log \\ rock} & cracks &  background \\
\hline
Side obstacles &  \makecell{fence \\ wall \\ pole \\ traffic sign \\ traffic light} &  \makecell{fence \\ container \\ table \\ pole \\ sign} & other & background \\
\hline
Water & - & water & water &  background \\
\hline
\end{tabular}
\caption{Mapping of dataset class labels to a common $L_{12}$ definition.}
\vspace{-5mm}
\label{tab:map}
\end{table}

To assess the impact of the training data environments, we train three SegFormer(MiT-B5) models independently on Cityscapes \cite{Cityscapes}, RUGD \cite{RUGD} and Earthquake \cite{earthquake}, considering the $L_{12}$ setting. The \textit{mIoU} results on the corresponding \textit{test}/\textit{val} sets are reported in Table~\ref{datasets}. As expected, each model achieves the highest \textit{mIoU} on its respective in-domain set, but exhibits significantly reduced performances on out-of-domain datasets. These large performance drops, up to $37.8$ p.p. on \textit{Cityscapes(val,$L_{12}$)}, $32.3$ p.p. on \textit{RUGD(test,$L_{12}$)}, and $32.0$ p.p. on \textit{Earthquake(test,$L_{12}$)}, highlight the substantial domain gaps between these datasets and the resulting limitations in cross-domain generalization.

We further evaluate the three models on the test and val splits of \benchm\, after performing the $L_{12} \rightarrow L_5$ mapping on the predictions such as detailed in Table~\ref{tab:map}. Figure~\ref{fig:predictions} illustrates qualitative results on images from the \benchm\ test set. The model trained on Cityscapes performs well in structured urban scenes (e.g., images 1 and 2), successfully segmenting classes omnipresent in such images such as vehicles and pedestrians. However, its performance degrades considerably in rural or damaged environments, where it struggles to differentiate between drivable and non-drivable areas and fails to identify road obstacles and blast craters (e.g., images 3–5). 
In contrast, the model trained on RUGD demonstrates better identification capacities of  road and drivable areas especially when confronted with less structured scenes compared to those from urban autonomous driving settings. Yet, it is less effective in detecting finer elements such as vehicles, pedestrians, and small obstacles. Meanwhile, the Earthquake-trained model yields the best segmentation results in destructed or post-disaster environments, particularly at detecting road obstacles, even the finer ones. However, it underperforms in recognizing vehicles and people due to their limited representation in the training data.

To leverage the strengths of each individual model, we train a SegFormer(MiT-B5) model on a combined dataset comprising Cityscapes, RUGD, and Earthquake, while maintaining the unified labelling strategy. This simple yet effective approach yields a model with strong and balanced perception capabilities across diverse outdoor environments: urban/rural, structured/destructed. Notably, it performs competitively on Cityscapes and RUGD compared to single-data models and achieves the best results on Earthquake, even surpassing the model trained solely on Earthquake data. Furthermore, it strongly outperforms all previous models on \benchm, as shown both quantitatively in Table~\ref{datasets} and qualitatively in Fig.~\ref{fig:predictions}. Thus, this model took advantage from Cityscapes for pedestrian and vehicle detection, has better separation abilities between road and drivable areas thanks to RUGD, and detects road obstacles, holes and debris learned thanks to Earthquake.

\section{Conclusion} 

In this work, we introduce \benchm, a new semantic segmentation benchmark \hejer{under data annotation frugality}, along with baseline evaluations to assess navigability in conflict-affected areas. Our approach begins with the construction of a dataset by filtering imagery from a publicly available DATTALION 
repository \cite{dattalion}. Then, we propose a refinement of the traditional mIoU metric to better reflect the requirements of autonomous vehicle navigation in unstructured environments. Subsequently, we benchmark several baselines on \benchm\ by varying architectures, backbones and training datasets without using any in-situ images during training. Building on these results, we propose a simple yet effective solution towards autonomous navigability in hazardous zones, leveraging the diversity of available annotated outdoor environments. 
Our experiments focus on direct transfer of models trained on other outdoor domains to compare baseline performances. A promising direction is to employ \benchm\ training dataset as a target domain and apply Unsupervised Domain Adaptation (UDA) techniques for semantic segmentation \cite{mic, attributes}, thereby improving model adaptation 
while remaining frugal in annotations. While our study provides initial insights and solutions to enhance unmanned vehicle safety in unstructured terrains, we believe UDA-driven approaches could further improve performance. Ultimately, we hope this work will foster research in such specific environments by providing open datasets and developing frugal and robust AI models.

\section{Broader Impact} 
\benchm\ represents a semantic segmentation dataset of war-affected environments, offering a first benchmark towards developing autonomous driving systems in such challenging domains. However, the methodologies used to construct this data introduce several important considerations that merit further investigation. First, the scraping of public multimedia repositories introduces potential vulnerabilities, such as the risk of malicious remote server image manipulation. Nonetheless, this approach significantly improves researcher safety by eliminating the need for data acquisition campaigns in active conflict zones. It also improves dataset representativeness when compared to artificially constructed environments, which may inadequately capture the complexity of real-world situations. Second, the use of images sourced from public areas raises compliance challenges with the GDPR when they contain identifiable individuals, including vulnerable populations. 
While autonomous vehicles are expected to process similar visual data in real time to avoid pedestrian collisions, the preparation, storage, and processing of corresponding training datasets requires explicit declaration and handling procedures under data protection regulations.





\section*{Acknowledgments}
Funded by the European Union. Views and opinions expressed are however those of the author(s) only and do not necessarily reflect those of the European Union or the European Commission. Neither the European Union nor the granting authority can be held responsible for them.
This work was supported by the European Union under the EDF Project FaRADAI (grant number 101103386).

This publication was made possible by the use of the FactoryIA supercomputer, financially supported by the Ile-De-France Regional Council. 

\bibliographystyle{IEEEtran}
\bibliography{Dattalion}

\end{document}